\documentclass[10pt,twocolumn,letterpaper]{article}

\usepackage[pagenumbers]{iccv} 
\usepackage{pifont}
\usepackage{amssymb}     
\usepackage{colortbl}       
\definecolor{mygray}{RGB}{240,240,240}
\usepackage{array}
\usepackage{multirow}
\usepackage{booktabs} 
\newcommand{\xmark}{\ding{55}}   
\newcommand{\cmark}{\ding{51}}   
\usepackage{makecell}
\definecolor{lightblue}{RGB}{217,235,255}  
\usepackage{minitoc}
\usepackage{graphicx}

\newcommand\blfootnote[1]{%
  \begingroup
  \renewcommand\thefootnote{}\footnote{#1}%
  \addtocounter{footnote}{-1}%
  \endgroup
}

\definecolor{iccvblue}{rgb}{0.21,0.49,0.74}
\usepackage[pagebackref,breaklinks,colorlinks,allcolors=iccvblue]{hyperref}

\def\paperID{4615} 
\def\confName{ICCV}
\def\confYear{2025}

\title{DriveX: Omni Scene Modeling for Learning Generalizable World Knowledge in Autonomous Driving}

\author{
Chen Shi$^{1{\diamond}}$ ~~~~Shaoshuai Shi$^{2}$ ~~~~ Kehua Sheng$^{2}$ ~~~~Bo Zhang$^{2}$ ~~~~Li Jiang$^{1\dagger}$ \\
{{$^1$}The Chinese University of Hong Kong, Shenzhen}\\
{{$^2$}Voyager Research, Didi Chuxing}\\
{\tt\small chenshi@link.cuhk.edu.cn~~jiangli@cuhk.edu.cn }\\
}

\begin{document}
\maketitle
\blfootnote{$\diamond$: Intern of Voyager Research; $\dagger$: Corresponding author.}

\begin{abstract}
Data-driven learning has advanced autonomous driving, yet task-specific models struggle with out-of-distribution scenarios due to their narrow optimization objectives and reliance on costly annotated data. We present DriveX, a self-supervised world model that learns generalizable scene dynamics and holistic representations (geometric, semantic, and motion) from large-scale driving videos. DriveX introduces Omni Scene Modeling (OSM), a module that unifies multimodal supervision—3D point cloud forecasting, 2D semantic representation, and image generation—to capture comprehensive scene evolution. To simplify learning complex dynamics, we propose a decoupled latent world modeling strategy that separates world representation learning from future state decoding, augmented by dynamic-aware ray sampling to enhance motion modeling.
For downstream adaptation, we design Future Spatial Attention (FSA), a unified paradigm that dynamically aggregates spatiotemporal features from DriveX’s predictions to enhance task-specific inference. 
Extensive experiments demonstrate DriveX’s effectiveness: it achieves significant improvements in 3D future point cloud prediction over prior work, while attaining state-of-the-art results on diverse tasks including occupancy prediction, flow estimation, and end-to-end driving. These results validate DriveX's capability as a general-purpose world model, paving the way for robust and unified autonomous driving frameworks.

\end{abstract}    
\section{Introduction}

\begin{figure}[t]
  \centering
   \includegraphics[width=0.999\linewidth]{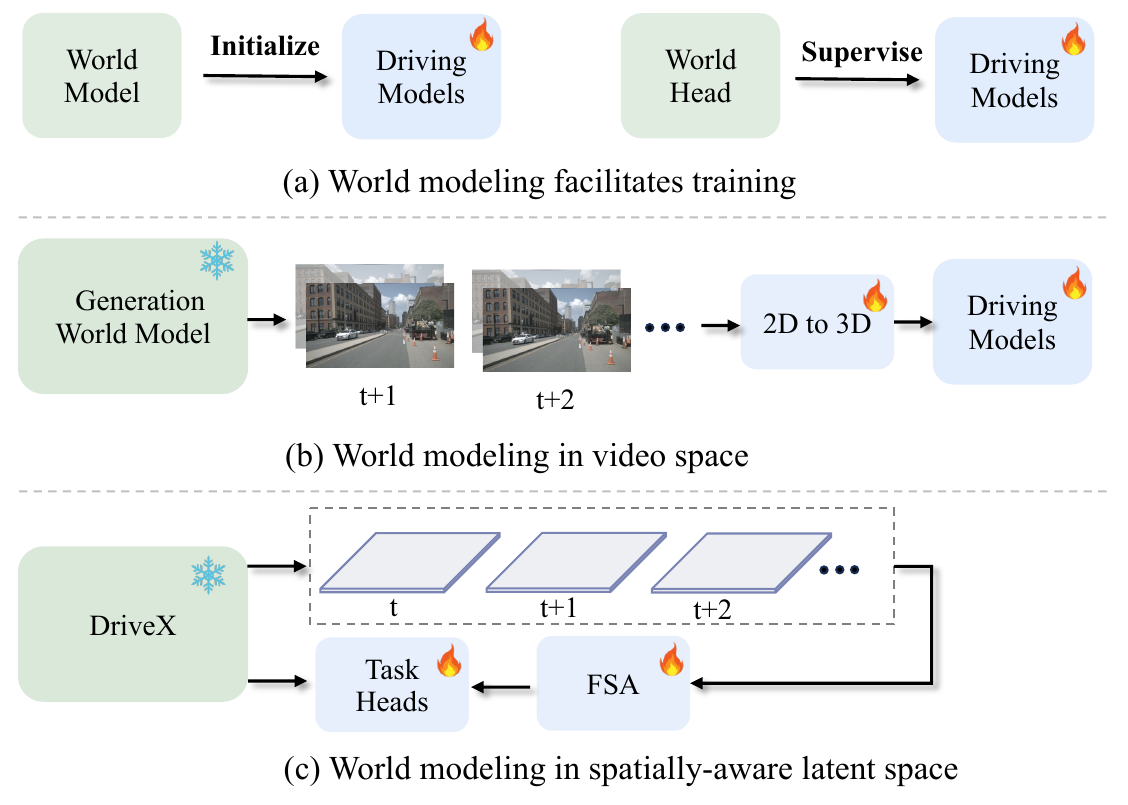}
   \vspace{-11pt}
   \caption{Comparison of different methods for integrating world models in autonomous driving systems. (a) World modeling serves as a pretraining task or an auxiliary supervision for driving models, without fully capturing general world knowledge. (b) Future prediction is performed in video space to assist downstream tasks, requiring additional 2D-to-3D transformation. (c) Our proposed DriveX framework decodes future states in latent 3D space, learning task-agnostic features that seamlessly benefit driving tasks. }
   \label{fig:intro}
   \vspace{-12pt}
\end{figure}

Data-driven learning techniques have significantly advanced autonomous driving, enhancing core tasks like perception~\cite{shi2023pv,lang2019pointpillars,yin2021center,wang2023exploring,zhang2023occformer} and planning~\cite{jiang2023vad,wang2024omnidrive,li2024ego,hu2023planning}. 
Despite this impressive progress, task-specific models still struggle with complex and out-of-distribution scenarios, primarily due to their task-specific optimization and reliance on limited human-annotated training data. 
World models, which predict future environment states based on historical observations and ego actions, have gained increasing attention for their ability to model complex driving environments~\cite{yang2024visual,min2024driveworld, yang2024driving,agro2024uno,guo2024end, wang2023drivedreamer,wang2024driving, gao2024vista,li2025drivingdiffusion}. By learning generalizable world knowledge from large-scale driving videos, world models hold the potential to improve robustness and safety across various autonomous driving tasks.

Existing world modeling approaches in autonomous driving can be broadly categorized into three strategies, as shown in Figure \ref{fig:intro}. The first strategy involves using world modeling as a general pretraining step~\cite{yang2024visual,min2024driveworld,agro2024uno}, followed by fine-tuning for specific driving tasks. The second strategy treats world modeling as an auxiliary head~\cite{10.1007/978-3-031-72624-8_4,yang2024driving,guo2024end,li2024enhancing,li2025navigationguided} during task-specific training. 
While effective, both strategies struggle to retain general world knowledge and may not fully exploit the world model's potential for detailed spatial and temporal prediction, as the model parameters are ultimately adapted for specific downstream tasks.
The third strategy uses world model as a generative model to predict future video sequences~\cite{wang2023drivedreamer,wang2024driving, gao2024vista,li2025drivingdiffusion,hu2023gaia} by utilizing video generation techniques~\cite{harvey2022flexible,khachatryan2023text2video,wang2024videocomposer}, which are subsequently used for task-specific purposes. Although this approach explicitly provides valuable predictive context by functioning as a world simulator, these generative models frequently struggle with maintaining consistent temporal-spatial coherence and lack adequate 3D spatial information, limiting their practical applications. These challenges motivate us to explore a fundamental question: how can we develop a world model framework that extracts generalizable world representations to benefit various downstream autonomous driving tasks within a unified paradigm?

We argue that an effective world model should excel in two key dimensions: 
1) it should perceive the environment by capturing both geometric structure and visual semantics—critical for ensuring that learned representations can be effectively transferred across diverse tasks; and 2) it should accurately model world dynamics, which is essential for decision-making tasks such as planning. In this work, we propose DriveX, a novel world model framework that not only preserves essential world knowledge within a spatially-aware bird’s-eye view (BEV) latent space—facilitating general knowledge transfer across multiple downstream tasks—but also captures precise world evolution. As illustrated in Figure \ref{fig:intro}(c), DriveX demonstrates its extensibility by seamlessly integrating with various autonomous driving tasks.

DriveX distinguishes itself from existing world modeling approaches through two key aspects. 
First, existing methods face a critical trade-off: prior work either focuses on geometric cues (e.g., point cloud forecasting~\cite{yang2024visual,zhang2024copilotd,khurana2023point}) or semantic occupancy prediction using manual labels~\cite{10.1007/978-3-031-72624-8_4, yang2024driving, DBLP:conf/cvpr/MinZ00XZ0LGXJN024}, sacrificing either visual details or scalable supervision.
To overcome these limitations, we introduce the Omni Scene Modeling (OSM) module, 
which encodes rich multimodal world knowledge into a bird’s-eye view (BEV) latent space.
Specifically, OSM integrates 3D point cloud forecasting for geometric fidelity, 2D semantic representations for class-aware reasoning, and image generation for retaining visual texture information.
As shown in Table~\ref{tab:intro}, when used as a pretrained feature extractor for end-to-end driving, DriveX maintains robust performance even with frozen parameters.
In contrast, ViDAR\cite{yang2024visual} and DriveWorld\cite{DBLP:conf/cvpr/MinZ00XZ0LGXJN024} experience significant performance degradation when parameters are frozen, highlighting the generalizable knowledge captured by our approach. Moreover, OSM operates in a fully self-supervised manner, leveraging either inherently available labels or those generated by foundation models~\cite{ren2024grounded,zhang2023simple} for scalable learning.

\begin{table}
\centering
\setlength{\tabcolsep}{11pt}  
\resizebox{0.9\linewidth}{!}{
\begin{tabular}{l|cc>{\columncolor[gray]{0.85}}c}
\toprule
Method & Finetuned & Fixed & Gap\\
\midrule
ViDAR~\cite{yang2024visual} & 83.6 & 77.0 & -6.6 \\
DriveWorld~\cite{DBLP:conf/cvpr/MinZ00XZ0LGXJN024}$\dagger$  & 83.7 &82.2& -1.5  \\
DriveX &83.7& 83.6 & -0.1  \\
\bottomrule
\end{tabular}
}
\vspace{-7pt}
\caption{End-to-end driving performance on NAVSIM~\cite{dauner2024navsim} test set. ``Finetuned'' refers to training the world model simultaneously, while ``Fixed'' represents freezing the world model. We observe negligible performance gap in DriveX. $\dagger$: re-implemented by us.}
\label{tab:intro}
\vspace{-16pt}
\end{table}

Second, we introduce a decoupled latent world modeling strategy that separates learning into two key components: world representation learning and latent future decoding. 
This hierarchical approach first builds a robust spatial representation before modeling temporal evolution, effectively capturing sparse motion signals.
In the world representation learning component, we train a world encoder to generate structured BEV features from multi-view historical images. These BEV features are optimized by our OSM module and thus are generalizable across driving tasks.  
In the latent future decoding component, we train a future decoder to model world evolution in the latent BEV space. Given the pretrained world encoder, the future decoder predicts motion flow and future BEV features based on given ego actions, with self-supervised alignment against the encoded future video frames by the world encoder. 
To further enhance motion modeling, we also apply OSM loss on the predicted latent features and introduce a dynamic-aware ray sampling strategy that prioritizes motion-salient regions.

To fully exploit DriveX’s predictive capabilities in downstream tasks, we introduce the Future Spatial Attention (FSA) paradigm, a lightweight yet effective module that seamlessly integrates DriveX into various driving applications. By acting as a ``world simulator'', DriveX predicts future latent BEV features over the next few seconds, providing a rich predictive context. FSA then employs task-specific queries to dynamically aggregate relevant contextual information from these features. This streamlined approach improves state-of-the-art performance across diverse tasks such as occupancy prediction, occupancy flow estimation, and end-to-end driving, highlighting DriveX's strong generalization and predictive capabilities.

In summary, our contributions are four-fold: 
1) We introduce DriveX, a self-supervised world model that learns generalizable world knowledge in latent space through Omni Scene Modeling with multimodal supervision; 
2) We propose a decoupled latent world modeling strategy that separates world representation learning from latent future decoding, enhanced by dynamic-aware ray sampling to prioritize motion-salient regions;
3) We present the Future Spatial Attention (FSA) paradigm, which leverages the pretrained world model to enhance diverse driving tasks within a unified framework; 
4) DriveX achieves significant improvements across multiple tasks, achieving a relative 6\% reduction in Chamfer Distance for 3-second future point cloud prediction, along with notable improvements in downstream tasks such as occupancy prediction, occupancy flow estimation, and end-to-end driving.

\section{Related Work}

\textbf{Point Cloud Forecasting} Point cloud forecasting, which aims to predict future point clouds from historical observations, has emerged as a crucial self-supervised learning paradigm in autonomous driving. Early approaches \cite{weng2021inverting, mersch2022self, weng2022s2net} typically project past point clouds into dense range images and leverage LSTM or 3D convolutional networks for future range image prediction. Recent work 4D-Occ \cite{khurana2023point} decomposes the task into 4D occupancy forecasting and employs differential depth rendering \cite{khurana2022differentiable,pan2024renderocc} to generate point cloud predictions from occupancy grids. COPILOT4D \cite{zhang2024copilotd} extends this line of work by incorporating diffusion models \cite{ho2020denoising, song2021denoising}. In a different direction, ViDAR \cite{yang2024visual} explores a vision-centric approach by predicting future point clouds from past visual inputs. Unlike existing researches that primarily utilize point cloud forecasting as a pretraining strategy for encoder in downstream tasks, our work adopts a fundamentally distinct perspective: we integrate point cloud forecasting as a component of the world model to learn more generalizable representations.

\noindent \textbf{World Models} World models \cite{ha2018recurrent,lecun2022path}, inspired by human cognition, enable intelligent agents to predict future outcomes based on past experiences and current actions. Learning world models has been widely studied to enhance policy function in reinforcement learning \cite{hafner2021mastering, hafner2023mastering, ha2018recurrent, hafner2019dream, sekar2020planning, wu2023daydreamer}, but their adoption in real-world applications remains limited. Another line of research emphasizes world simulation, aiming to generate high-fidelity future videos in gaming~\cite{alonso2024diffusion,che2024gamegen,valevski2024diffusion} and real-world scenarios~\cite{CogVideo, CogVideoX, T2V-Turbo, chen2024videocrafter2}. However, these methods often fail to provide consistent and precise spatiotemporal representation. In this work, we target real-world autonomous driving and construct world models directly in the spatial BEV space, enabling transferable representations that support diverse downstream driving tasks.

\noindent \textbf{World Modeling for Autonomous Driving} As a complex and safety-critical real-world application, autonomous driving has increasingly attracted interest in world modeling. Leveraging advances in video generation, several approaches \cite{wang2023drivedreamer,wang2024driving, gao2024vista,li2025drivingdiffusion,gao2023magicdrive,gao2024magicdrive3d,hu2023gaia} have demonstrated the capability to synthesize realistic driving sequences across diverse scenarios. However, the generated videos lack comprehensive 3D scene understanding, restricting their applications in downstream tasks. Alternative methods \cite{10.1007/978-3-031-72624-8_4, min2024driveworld, yang2024driving, li2025semisupervised} explore future prediction in occupancy space, achieving remarkable progress but relying on costly semantic occupancy annotations. Meanwhile, research on applying world models to real-world driving applications mainly focuses on using world models as auxiliary supervision \cite{yang2024visual,li2024enhancing, li2024enhancing,li2025navigationguided} for autonomous model training, failing to capture task-agnostic world features. Recent works such as LAW~\cite{li2024enhancing} and SSR~\cite{li2025navigationguided} have also explored world modeling in latent BEV space to eliminate the need for manual labels, but are limited to the end-to-end driving task. To address these challenges, we present DriveX, a spatially-aware world model that learns general world knowledge in a self-supervised manner and offers seamless integration with various driving tasks via spatial attention. 
\begin{figure*}[!t]
  \centering
  \includegraphics[width=.99\linewidth]{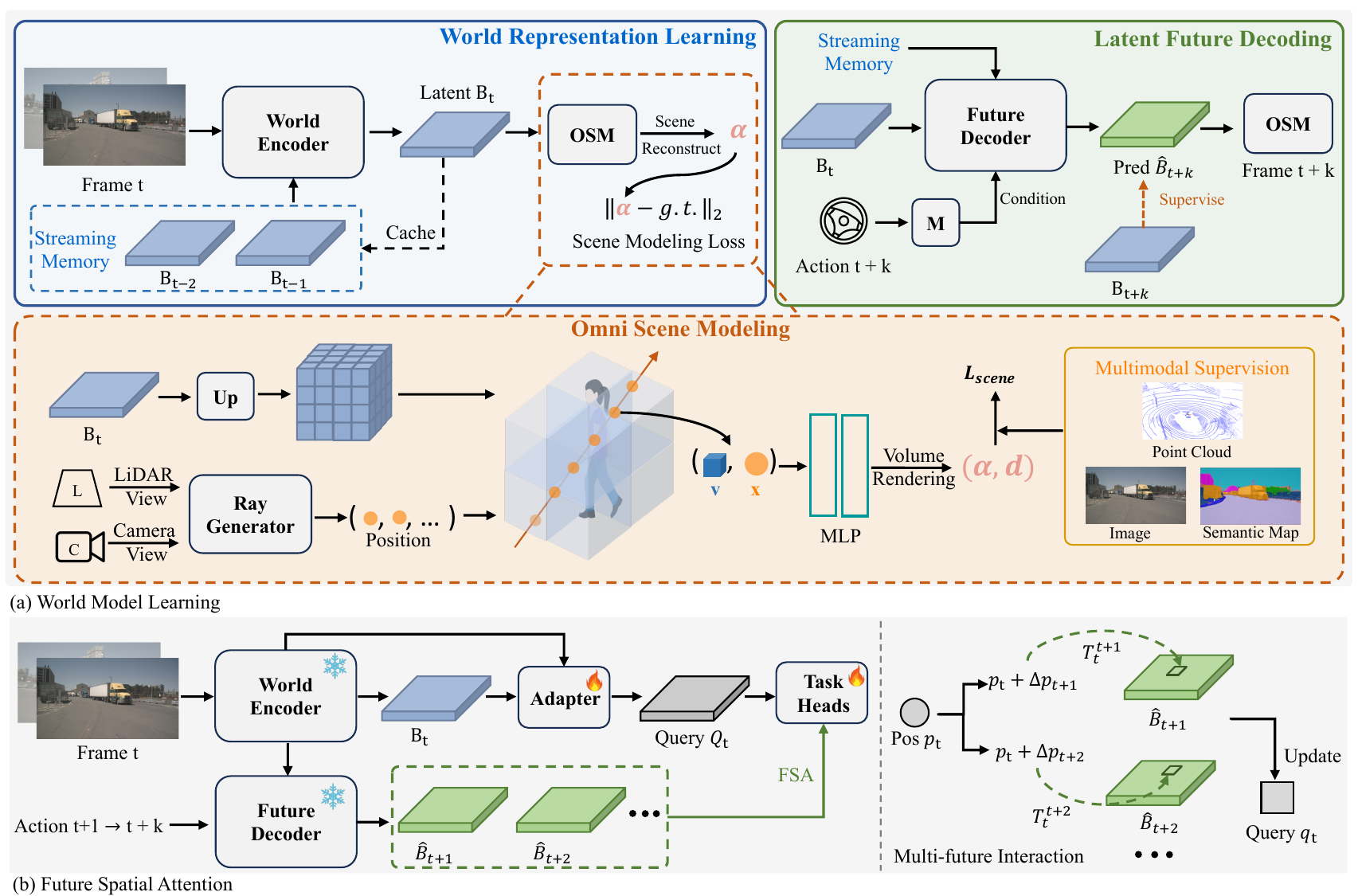}
  \vspace{-8pt}
  \caption{\textbf{Top:} An overview of our DriveX framework. The learning process consists of two stages: world representation learning, where the model learns temporal and geometric semantics through Omni Scene Modeling, and latent future decoding, where the model predicts future states in the learned latent space. Both stages are trained in a self-supervised manner. \textbf{Bottom:} Illustration of the FSA paradigm. Various driving tasks can dynamically aggregate information from predicted latent features through FSA. }
  \label{fig:method}
  \vspace{-5pt}
\end{figure*}

\section{Methodology}

In autonomous driving scenarios, world models are typically designed to predict future environmental states from past observations and ego actions, enabling generalized scene understanding. Some approaches leverage world modeling as a pre-training step~\cite{yang2024visual,zhang2024copilotd,khurana2023point} or as an auxiliary task~\cite{harvey2022flexible,khachatryan2023text2video,wang2024videocomposer} but often compromise transferability due to reliance on specialized supervision. 
Another line of world models~\cite{wang2023drivedreamer,wang2024driving,gao2024vista,li2025drivingdiffusion} explores to generate future video sequences but faces challenges in ensuring the spatial and temporal consistency.

To advance the utility of the world model in driving tasks, we propose DriveX, as depicted in Figure~\ref{fig:method}. DriveX leverages multi-view historical camera video and ego action data to predict future world states within a spatially-aware latent space. To learn general world representation, we propose the Omni Scene Modeling module (described in Sec~\ref{chapter:method-encoder}), which employs a multimodal supervision strategy to simultaneously encode geometric, semantic, and visual details in a self-supervised manner. To further improve world dynamics modeling, we design a decoupled learning strategy that enables precise motion signal learning based on well-trained BEV latent representations (See Sec~\ref{chapter:method-decoder}). Finally, we present the Future Spatial Attention (FSA) mechanism (See Sec~\ref{chapter:method-downstream}), which seamlessly integrates DriveX with various downstream driving tasks.

\subsection{World Representation Learning} \label{chapter:method-encoder}
To encode environmental information into latent representations, prior works have explored point cloud reconstruction~\cite{yang2024visual,khurana2023point} and occupancy prediction as proxy tasks~\cite{10.1007/978-3-031-72624-8_4, yang2024driving, DBLP:conf/cvpr/MinZ00XZ0LGXJN024}. However, these specialized objectives fail to fully capture comprehensive scene information (as demonstrated in Table~\ref{tab:intro}). To address this limitation, we design the Omni Scene Modeling (OSM) module that promotes the learning of generalizable representations. Leveraging foundation models~\cite{ren2024grounded,zhang2023simple} and paired multi-view images and point cloud data, this learning process is fully self-supervised.

\vspace{1mm}
\noindent\textbf{Temporal Streaming World Encoder.} 
Given a sequence of multi-view images $I_{1:t}$ up to the current timestamp $t$, we apply a world encoder to compress the multi-view historical observations into a latent BEV representation $B_{t} \in \mathbb{R}^{H \times W \times C}$, where $H$, $W$ and $C$ are the width, height and number of channels of the feature map, respectively. We follow the BEVFormer \cite{li2022bevformer} paradigm for 2D-to-3D view transformation. Taking previous latent BEV features $B_{t-1}$ in streaming memory and images $I_t$ from $N_{\text{view}}$ cameras as inputs, the world encoder extracts visual information and transforms it into grid-structured BEV features, as follows:
\begin{equation} \label{equ:world_encoder}
\vspace{-1pt}
    B_t = \operatorname{TCA}(E(I_t), B_{t-1}),
\vspace{-1pt}
\end{equation}
where $E$ denotes the combination of a 2D encoder \cite{he2016deep,lin2017feature,wang2023internimage} and a 2D-to-3D transformation network \cite{li2022bevformer}. TCA represents a Temporal Cross-Attention module \cite{zhu2020deformable} that achieves interaction between current states and historical latent information. The encoder operates in a streaming manner, continuously caching BEV features in the streaming memory to maintain long-term temporal dependencies.

In this way, we obtain a compact and structured representation $B_t$, where each grid cell corresponds to a specific location in the physical world. The spatial priors of grids in $B_t$ facilitate scene dynamics modeling by enabling independent processing of each grid cell. Furthermore, such structured representation ensures strong compatibility with diverse practical applications (detailed in Section \ref{chapter:method-downstream}).

\vspace{1mm}
\noindent\textbf{Omni Scene Modeling.} 
A well-designed training signal is essential for the world encoder to capture generalizable latent BEV features. To achieve this, we design a omni scene modeling module to compel the world encoder to develop a comprehensive understanding of the environment. Specifically, we categorize the environmental understanding into three aspects: geometric structure, semantic information, and visual details. Then we introduce a multimodal supervision strategy, combining 3D point cloud reconstruction, 2D semantic prediction, and image generation to improve the richness of the latent representation.

To be specific, let $I_t \in\mathbb{R}^{N_{\text{view}}\times H_\text{img} \times W_\text{img} \times 3} $ and $P_t \in \mathbb{R}^{N_{p} \times 3} $ represent the multi-view images with shape ($W_\text{img}$, $H_\text{img}$) and point clouds with $N_p$ points, respectively. Firstly, we cast $N_r$ rays from image and LiDAR views by sampling corresponding pixels and points. Each ray can be parameterized as $r_i = o_i + t d_i$, where $o_i$ denotes the camera or LiDAR center and $d_i$ indicates the ray direction pointing to the sampled pixel or point. Then we sample $n$ waypoints $x_i \in \mathbb{R}^{n \times 3}$ along the ray $r_i$. The coordinates of these waypoints are computed as follows:
\begin{equation}
\vspace{-1pt}
    x_{i,j} = o_i + \lambda_{j} d_i,
\vspace{-1pt}
\end{equation}
where $i$ and $j$ denotes the index of ray and waypoint, respectively, and $\lambda=\{\lambda_1,\lambda_2, ..., \lambda_n\}$ represents predefined distances.

The waypoint features $V_t \in \mathbb{R}^{N_r \times n \times C^*}$ are obtained by first transforming the BEV features $B_t$ into 3D space through a Channel-to-Height module, yielding volumetric features $F_t \in \mathbb{R}^{H \times W \times Z \times C^*}$, followed by a trilinear interpolation operation:
\begin{equation} \label{eq:occ_sample}
\vspace{-1pt}
\begin{split}
F_t &= \operatorname{Reshape}(\operatorname{Conv}(B_t), [H,W, Z, C^*]), \\
\quad V_t &= \operatorname{Interp}(F_t, X_t),
\end{split}
\vspace{-1pt}
\end{equation}
where $\operatorname{Conv}$ denotes 2D convolution layer with input channels $C$ and output channels $ZC^*$, and $X_t =\{x_i\}_{i=1}^{N_r}$ represents the collection of ray waypoints. Subsequently, for each waypoint $x_{i,j}$, we utilize a MLP to predict its attribution filed $\alpha_{i,j}\in \mathbb{R}^{N_s + N_c}$ and density field $\sigma_{i,j} \in \mathbb{R} $:
\begin{equation}
\vspace{-1pt}
    (\alpha_{i,j}, \sigma_{i,j})= \text{MLP}(\text{Concat}(v_{i,j}, \text{PE}(x_{i,j})), 
\vspace{-1pt}
\end{equation}
where $v_{i,j}$ represents the waypoint features from $V_t$, $N_s$ and $N_c$ indicate the number of semantic classes and RGB color channels, respectively. $\text{PE}$ is the cosine position encoding. Finally, the attribute prediction $\alpha_i$ for $r_i$ is accumulated through volume rendering process~ \cite{pan2024renderocc,yang2024pred}:
\begin{equation} \label{eq:render}
\vspace{-1pt}
\begin{split}
    \tau_j &= \exp(-\sum_{l=1}^{j-1} \sigma_{i,l}\delta_l), \\
    \alpha_i &= \sum_{j=1}^{n} \tau_j (1-\exp(-\sigma_{i,j}\delta_j)) \alpha_{i,j},
\end{split}
\vspace{-1pt}
\end{equation}
where $\delta_j = \lambda_{j} - \lambda_{j-1}~~(\lambda_0 = o_i)$ represents the distance between two adjacent waypoints, and $\alpha_i$ can be decomposed into semantic prediction $s_i \in \mathbb{R}^{N_s}$ and color prediction $c_i \in \mathbb{R}^{N_c}$. Following the same formulation, the expected depth $d_i$ of $r_i$ is estimated by replacing $\alpha_{i,j}$ with the distance $\lambda_j$ in Eq.(\ref{eq:render}).

To enable comprehensive scene understanding, we train the world encoder using a multimodal supervision framework, deriving the scene modeling loss $\mathcal{L}_\text{scene}$ from both camera and LiDAR perspectives. Specifically, for camera-view rays, we optimize the scene modeling loss $\mathcal{L}_\text{scene}^{\text{camera}}$ that integrates semantic segmentation ($\mathcal{L}_\text{sem}$) and color reconstruction ($\mathcal{L}_\text{rgb}$) objectives. For LiDAR-view rays, we optimize the scene modeling loss $\mathcal{L}_\text{scene}^{\text{LiDAR}}$ with only depth estimation ($\mathcal{L}_\text{depth}$):
\begin{equation} 
\vspace{-1pt}
\begin{split}
    \mathcal{L}_\text{scene}^{\text{camera}} &= \mathcal{L}_\text{sem} + \mathcal{L}_\text{rgb},\\
    \mathcal{L}_\text{scene}^{\text{LiDAR}}  &= \mathcal{L}_\text{depth},
\end{split}
\vspace{-1pt}
\end{equation}
where we adopt cross-entropy loss for $\mathcal{L}_\text{sem}$ and L2 Loss for both $\mathcal{L}_\text{rgb}$ and $\mathcal{L}_\text{depth}$. The ground truth depth and color values are directly acquired from raw sensor data, while semantic labels are automatically generated using foundation models Grounded SAM~\cite{ren2024grounded} and OpenSeeD~\cite{zhang2023simple}, enabling scalable training without manual annotation.

\begin{figure}[t]
  \centering
   \includegraphics[width=0.9\linewidth]{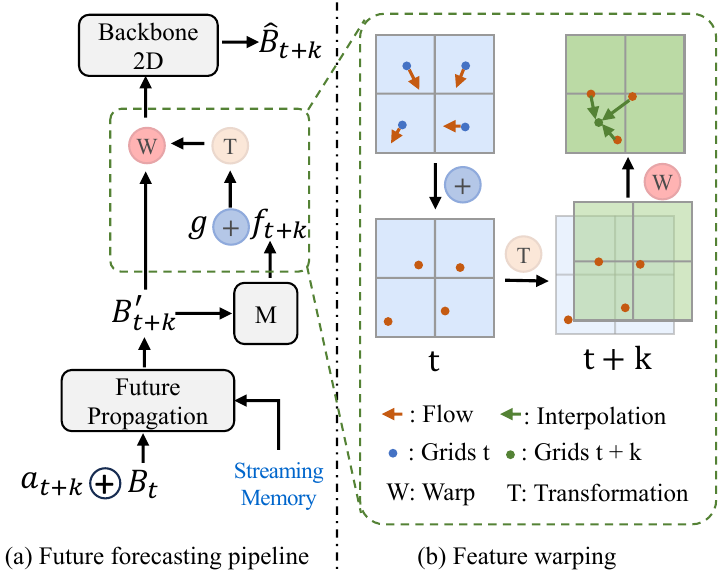}
   \vspace{-6pt}
   \caption{Illustration of the flow-based future forecasting pipeline. (a) The detailed structures of future decoder, consisting of a future propagation module, a motion head ``M'', and a refinement 2D convolutional backbone. (b) Grid points with predicted flows are transformed to time $t+k$ coordinates, followed by distance-based interpolation to obtain the future grid features.}
   \label{fig:flow}
   \vspace{-12pt}
\end{figure}

\subsection{Decoupled Latent World Modeling}\label{chapter:method-decoder}
Instead of jointly learning representation and temporal dynamics, we propose a decoupled strategy, where the world decoder is first trained to generate representative BEV features that reflect a comprehensive understanding of past observations. Building on these spatially-aware features, we then introduce a dynamic future decoder module to model temporal evolution in the latent BEV space, without being burdened by feature learning. Specifically, we employ an explicit flow-based future forecasting strategy to generate future world states in the latent BEV space, along with a dynamic-aware ray sampling strategy to improve learning efficiency in potentially dynamic regions.

\noindent\textbf{Flow-based Future Forecasting.} As illustrated in Figure \ref{fig:flow}, the future decoder predicts future states based on the latent spatial features $B_t$ and action condition, \textit{i.e.}, ego-motion. Specifically, the ego-motion to target time $t+k$ is first encoded into an embedding $a_{t+k} \in \mathbb{R}^{C}$ through a multi-layer perceptron (MLP), which is added to $B_t$ to incorporate action information. Then, a future propagation module consisting of stacked deformable attention blocks processes these condition-augmented features and interacts with historical latent features in streaming memory, producing future features $B_{t+k}^{\prime}$. The corresponding future grid positions 
$g_f$ are computed by adding current grid positions $g \in \mathbb{R}^{H \times W \times 2}$ with flow fields $f_{t+k} \in \mathbb{R}^{H \times W \times 2}$, where $f_{t+k}$ is predicted from $B_{t+k}^{\prime}$ via a motion head. Since $g_f$ are expressed in the frame-$t$ coordinate system, we transform them to the target frame $t+k$ as follows:
\begin{equation}
\vspace{-1pt}
    g_{f}' = {T}_{t}^{t+k} g_{f},
\vspace{-1pt}
\end{equation}
where ${T}_{t}^{t+k}$ is the ego-motion transformation matrix from the current frame to frame $t+k$. Finally, distance-based interpolation is applied to warp the features into grid-structure future state ${\hat B}_{t+k}$. In particular, for each grid in ${\hat B}_{t+k}$, we select $N$ nearest neighbor grids in $B_{t+k}^{\prime}$ using $g_{f}'$, 
and aggregate the grid features through inverse distance weighted summation. A 2D convolutional backbone is employed to further refine the interpolated features. During training, we supervise the predicted features $\hat{B}_{t+k}$ using ground truth latent representations ${B}_{t+k}$ extracted by the world encoder:
\begin{equation}
\vspace{-1pt}
    \mathcal{L}_{\text{latent}} = \Vert{B}_{t+k} -\hat{{B}}_{t+k}\Vert.
\vspace{-1pt}
\end{equation}
In this way, we can predict future states at arbitrary timestamps. Compared to autoregressive methods~\cite{yang2024visual,10.1007/978-3-031-72624-8_4}, this direct prediction strategy not only avoids error accumulation but also offers greater flexibility in practice.

\noindent \textbf{Dynamic-aware Ray Sampling.} In addition to latent space supervision, we incorporate scene modeling loss $\mathcal{L}_{\text{scene}}$ at future timestamps. However, since the majority of rays belong to static background elements, salient moving objects may be overlooked during training. To bolster the model's understanding of world dynamics, we present a dynamic-aware ray sampling strategy, which leverages an off-the-shelf tracker to identify regions of interest (RoIs) with high dynamic activity, preserving rays within these regions to prioritize motion-salient areas.

\noindent \textbf{Learning Objective} To summarize, the overall objective for the future decoder contains latent space alignment and future scene modeling:
\begin{equation} \label{eq:dynamic_loss}
\vspace{-1pt}
    \mathcal{L} _{\text{future}} = \sum_{k=1}^{F} \left(\omega_l \mathcal{L} _{\text{latent}}^k + \omega_s \mathcal{L} _{\text{scene}}^k \right),
\vspace{-1pt}
\end{equation}
where $\omega_l$ and $\omega_s$ are the weight coefficients, and $F$ represents the number of future timestamps used for supervision. Notably, the supervision signal ${B}_{t+k}$ in $\mathcal{L}_{\text{latent}}$ is generated from raw images using Eq.(\ref{equ:world_encoder}), while $\mathcal{L} _{\text{scene}}$ operates without manual annotations, ensuring a fully self-supervised training paradigm in future decoding.

\subsection{World Model Integration with Task Heads}\label{chapter:method-downstream}
The fully-trained DriveX model can predict future world evolutions in the latent BEV space, offering valuable insights for a range of autonomous driving tasks. Notably, the generated generic and spatially-aware BEV features are readily compatible with various driving models. As shown in the bottom part of Figure~\ref{fig:method}, various downstream tasks can directly employ the BEV features from the frozen world model through a lightweight adapter, which enhances $B_t$ with latent image features from world encoder using a stack of deformable attention layers. Then, we introduce a Future Spatial Attention (FSA) paradigm that enables future predictions to be incorporated into various autonomous driving models within a unified framework. To illustrate the integration of the DriveX model, we focus on two representative tasks: occupancy prediction and end-to-end driving.

\noindent \textbf{Action Prediction.} 
Because the world model's future predictions are conditional on the ego vehicle's future actions, we employ a straightforward strategy to predict these actions for models that don't inherently include this functionality, such as perception models.
Specifically, we follow AD-MLP \cite{zhai2023rethinking} to predict actions based on ego states and historical trajectories, focusing on short-term predictions to simplify the task compared to full trajectory planning. For the end-to-end driving task, we directly adopt the planning head's output as the predicted actions.

\noindent \textbf{Future Spatial Attention Paradigm.} 
To integrate the predicted future BEV features into downstream tasks, we propose a unified Future Spatial Attention (FSA) paradigm. As shown in Figure~\ref{fig:method}, this paradigm uses task-specific queries to directly aggregate world state information from the predicted future BEV features via standard spatial attention. Concretely, given a query $q$ from the task model and its 3D reference point $p$, the spatial attention can be formulated as:
\begin{equation}
\vspace{-1pt}
    q := q + \sum_{k=1}^{K} \sum_{j=1}^{J} A_{kj}W {\hat{B}}_{t+k}[{T}_{t}^{t+k}(p + \Delta p_{kj})],
\vspace{-1pt}
\end{equation}
where $K$ and $J$ denote the number of future timesteps and sampling positions, respectively. $W$ represents the learnable projection matrix, $A_{kj}$ indicates the normalized attention weight, and $\Delta{p}_{kj}$ is sample offset. Unlike conventional deformable attention, we transform the sample point $p + \Delta p_{kj}$ to the frame-$(t\!+\!k)$ coordinate system for feature sampling from ${\hat{B}}_{t+k}$. For simplicity, we omit the explicit expression of attention heads. The key to adapting spatial attention for various downstream tasks lies in the selection of $\Delta{p}_{kj}$, which enables the DriveX model to seamlessly integrate with diverse autonomous driving tasks while preserving the integrity of existing architectures.

\begin{table}
\centering
\resizebox{\columnwidth}{!}{
\begin{tabular}{c|c|cccccc}
\toprule
\multirow{2}{*}{Method} & \multirow{2}{*}{Modality} & \multicolumn{6}{c}{Chamfer Distance ($\text{m}^2$) $\downarrow$} \\
& & 0.5s & 1.0s & 1.5s & 2.0s & 2.5s & 3.0s \\
\midrule
4D-Occ${}^*$ \cite{khurana2023point} & L & 0.91 & 1.13 & 1.30 & 1.53 & 1.72 & 2.11 \\
ViDAR~\cite{yang2024visual} & C & 1.01 & 1.12 & 1.25 & 1.38 & 1.54 & 1.73 \\
HERMES \cite{zhou2025hermes} & C & - & 0.78 & - & 0.95 & - & 1.17 \\
DriveX-B & C & \textbf{0.55} & \textbf{0.66} & \textbf{0.75} & \textbf{0.86} & \textbf{0.97} & \textbf{1.10} \\
\bottomrule
\end{tabular}
}
\vspace{-9pt}
\caption{Point cloud forecasting performance comparison on nuScenes \cite{caesar2020nuscenes} dataset. L and C denote LiDAR and Camera inputs, respectively. $*$: re-evaluated by \cite{yang2024driving}.}
\label{tab:point}
\vspace{-12pt}
\end{table}
\section{Experiments}

\begin{table*}[!t]
	\setlength{\tabcolsep}{0.0035\linewidth}
	\newcommand{\classfreq}[1]{{~\tiny(\semkitfreq{#1}\%)}}  %
	\centering
   \resizebox{1\linewidth}{!}{
	\begin{tabular}{l|c|c|ccc| c c c c c c c c c c c c c c c c c}
 
		\toprule
		Method
        & \makecell[c]{Backbone}
		& \makecell[c]{Flow\\Head}
		& \makecell[c]{\colorbox[gray]{0.85}{mIoU}}
            & \makecell[c]{$\text{IoU}_{geo}$}
            & \makecell[c]{\colorbox[gray]{0.85}{$\text{mAVE}\downarrow$}}
            & \rotatebox{90}{others} 
		& \rotatebox{90}{barrier} 
		& \rotatebox{90}{bicycle}
		& \rotatebox{90}{bus} 
		& \rotatebox{90}{car} 
		& \rotatebox{90}{const. veh.} 
		& \rotatebox{90}{motorcycle} 
		& \rotatebox{90}{pedestrian} 
		& \rotatebox{90}{traffic cone} 
		& \rotatebox{90}{trailer} 
		& \rotatebox{90}{truck} 
		& \rotatebox{90}{drive. suf.} 
		& \rotatebox{90}{other flat} 
		& \rotatebox{90}{sidewalk} 
		& \rotatebox{90}{terrain} 
		& \rotatebox{90}{manmade} 
		& \rotatebox{90}{vegetation} \\
		\midrule

            BEVFormer\cite{li2022bevformer} & InternT & \xmark &36.93 & 68.49 & - & 8.56 & 42.94 & 19.34 & 47.02 & 49.59 & 20.36 & 22.62 & 24.69 & 19.77 & 30.11 & 35.24 & 82.11 & 40.86 & 50.62 & 54.81 & 42.56 & 36.55  \\
            CVT-Occ \cite{ye2024cvt}& Res101 & \xmark & 40.34 & - & - & 9.45 & 49.46 & 23.57 & 49.18 & 55.36 & 23.10 & 27.85 & 28.88 & 29.07 & 34.97 & 40.98 & 81.44 & 40.92 & 51.37 & 54.25 & 45.94 & 39.71  \\
            ViewFormer \cite{li2025viewformer} & InternT & \xmark & 43.61 & 72.46 & - & 13.82 & 50.32 & 29.49 & 49.24 & 54.52 & 24.34 & 32.72 & 31.09 & 31.49 & 34.44 & 41.62 & 85.47 & 51.27 & 59.03 & 62.15 & 48.33 & 42.06 \\
            FB-OCC \cite{li2023fb} & InternT & \xmark &38.69 & 69.95 & - & 11.40 & 41.42 & 24.27 & 46.01 & 49.38 & 24.56 & 27.06 & 28.09 & 25.61 & 32.23 & 38.46 & 80.97 & 42.99 & 50.95 & 56.15 & 40.55 & 37.61  \\
            ViewFormer (ViDAR \cite{yang2024visual})$\dagger$ & InternT & \xmark & 43.72 & 72.88 & - & 13.52 & 50.69 & 29.56 & 49.77 & 54.72 & 25.24 & 32.41 & 31.59 & 31.20 & 34.54 & 42.03 & 85.40 & 51.01 & 58.96 & \textbf{62.28} & 48.10 & 42.17 \\
            ViewFormer \cite{li2025viewformer} & InternT & \xmark & 43.61 & 72.46 & - & 13.82 & 50.32 & 29.49 & 49.24 & 54.52 & 24.34 & 32.72 & 31.09 & 31.49 & 34.44 & 41.62 & 85.47 & 51.27 & 59.03 & 62.15 & 48.33 & 42.06 \\
            ViewFormer (DriveX-S) & InternT & \xmark &  \textbf{44.38} &  \textbf{73.61} &  -  & \textbf{14.00} & \textbf{51.18} & \textbf{30.10} & \textbf{50.13} & \textbf{55.44} & \textbf{26.82} & \textbf{33.12} & \textbf{32.37} & \textbf{32.07} & \textbf{35.60} & \textbf{43.18} & \textbf{85.78} & 50.93 & \textbf{59.16} & 62.23 & \textbf{49.80} & \textbf{42.59} \\
            \rowcolor{lightblue}
            \textit{Improvement} &  &  &  +\textbf{\textit{0.77}} &  +\textbf{\textit{1.15}} &  & & & & & & & & &
            & & & & & & & &\\

            \midrule

            BEVFormer${}^*$  \cite{li2022bevformer} & InternT & \cmark &33.61 & 67.41 & 0.695 & 8.44 & 40.80 & 12.57 & 37.70 & 45.76 & 18.44 & 12.14 & 22.52 & 21.25 & 25.97 & 29.40 & 80.92 & 38.19 & 48.56 & 52.58 & 40.98 & 35.07  \\
            FB-OCC${}^*$  \cite{li2023fb} & InternT & \cmark &37.36 & 69.73 & 0.433 & 10.95 & 40.08 & 23.14 & 42.87 & 47.17 & 21.43 & 23.84 & 27.24 & 24.50 & 31.54 & 37.36 & 80.31 & 42.26 & 50.14 & 55.44 & 39.98 & 36.84  \\
            ViewFormer (ViDAR~\cite{yang2024visual})$\dagger$ & InternT & \cmark & 42.64 & 72.76 & 0.404 & \textbf{13.73} & 49.23 & 29.04 & 46.98 & 52.79 & 21.80 & 29.85 & 30.73 & 30.55 & 34.42 & \textbf{41.22} & 85.17 & 50.45 & 58.28 & 61.09 & 47.82 & 41.72 \\
            ViewFormer \cite{li2025viewformer} & InternT & \cmark & 42.54 & 72.36 & 0.412 & 13.63 & 49.35 & 28.74 & 46.63 & 52.71 & 21.04 & 29.63 & 30.34 & 30.53 & 34.01 & 41.04 & 85.23 & 50.63 & 58.68 & 61.63 & 47.72 & 41.56 \\
            ViewFormer (DriveX-S) & InternT & \cmark &  \textbf{43.47} &  \textbf{73.44} &  \textbf{0.385}  & 13.72 & \textbf{49.46} & \textbf{30.64} & \textbf{48.85} & \textbf{54.99} & \textbf{27.13} & \textbf{31.90} & \textbf{32.62} & \textbf{31.29} & \textbf{34.29} & 41.16 & 84.89 & 49.91 & 58.00 & 60.75 & 47.50 & \textbf{41.88}  \\
            \rowcolor{lightblue}
            \textit{Improvement} &  &  &  +\textbf{\textit{0.93}} &  +\textbf{\textit{1.08}} &  -\textbf{\textit{0.027}}  & & & & & & & & &
            & & & & & & & &\\
            
		\bottomrule
	\end{tabular}
     }
\vspace{-9pt}
 \caption{3D occupancy and occupancy flow prediction performance comparison on FlowOcc3D \cite{li2025viewformer}. We report results for models with and without a flow head. Performance gains achieved by our DriveX model are highlighted. ${}^*$: reported by \cite{li2025viewformer}. $\dagger$: re-implemented by us.}
 \label{tab:occ}
 \vspace{-10pt}
\end{table*}

\begin{table}
\centering
\resizebox{\linewidth}{!}{
\setlength{\tabcolsep}{2pt}
\begin{tabular}{l|c|cc|ccc|>{\columncolor[gray]{0.85}}c}
\toprule
Method & Modality & NC $\uparrow$ & DAC $\uparrow$ & TTC $\uparrow$ & Comf. $\uparrow$ & EP $\uparrow$ & PDMS $\uparrow$\\
\midrule
Constant Velocity~\cite{dauner2024navsim} & Camera  & 68.0 & 57.8 & 50.0 & 100 & 19.4 &  20.6  \\
Ego Status MLP~\cite{dauner2024navsim}& Camera   & 93.0 & 77.3 & 83.6 & 100 & 62.8 & 65.6  \\
Transfuser${}^*$~\cite{chitta2022transfuser}& Camera  & 97.4 & 92.8 & 92.4 & 100 & 79.0 & 83.8  \\
PARA-Drive ${}^*$~\cite{weng2024drive}& Camera  & 97.9 & 92.4 & 93.0 & 99.8 & 79.3 & 84.0  \\
UniAD${}^*$~\cite{hu2023planning}& Camera   & 97.8 & 91.9 & 92.9 & 100 & 78.8 & 83.4  \\
\hline
DiffusionDrive-C~\cite{liao2024diffusiondrive}$\dag$& Camera    & 97.8 & 92.2 & 92.6 & 99.9 & 78.9 & 83.6  \\
DriveX-S& Camera    & 97.5 & \textbf{94.0} & \textbf{93.0} & \textbf{100} & \textbf{79.7} & \textbf{84.5}  \\
\rowcolor{lightblue}
\textit{Improvement} & - & \textit{-0.3} & \textbf{\textit{+1.8}}  & \textbf{\textit{+0.4}} & \textbf{\textit{+0.1}} & \textbf{\textit{+0.8}} &  \textbf{\textit{+0.9}} \\
\bottomrule
\end{tabular}
}
\vspace{-9pt}
\caption{End-to-end planning performance on NAVSIM test split. ``PDMS'' indicates the predictive driver model score. ${}^*$: reported by \cite{dauner2024navsim}. $\dag$: our re-implementation of the camera-based DiffusionDrive model. The best results are marked in \textbf{bold}.}
\label{tab:e2e}
\vspace{-12pt}
\end{table}

\subsection{Experimental Settings}
\textbf{Dataset.} We conduct experiments on two multimodal autonomous driving datasets: nuScenes \cite{caesar2020nuscenes} and NAVSIM~\cite{dauner2024navsim}. The nuScenes dataset comprises 1,000 driving sequences and supports various autonomous driving tasks. For occupancy prediction and occupancy flow estimation experiments, we utilize the annotations from the FlowOcc3D benchmark~\cite{li2025viewformer}. The NAVSIM dataset, built upon OpenScene~\cite{openscene2023}, provides 115k filtered samples for challenging end-to-end driving. We generate image semantic labels for both datasets using 2D foundation models~\cite{ren2024grounded,zhang2023simple}, resulting in 15 classes as defined in \cite{zhang2023occnerf}. 

\noindent\textbf{Network Architecture.} The world encoder consists of an InternImage-Tiny \cite{wang2023internimage} backbone with an FPN neck~\cite{lin2017feature}, followed by six encoder layers~\cite{li2022bevformer} to extract BEV-space world representations from multi-view images and historical latent states. The future decoder consists of a future propagation module with three deformable attention blocks \cite{li2022bevformer}, a motion head, and a 2D CNN backbone for refinement. We develop two variants of the world model: DriveX-S, which uses a smaller input image size of $256\times704$, and DriveX-B, which employs a larger size of $900\times1600$. For downstream task verifications, we adopt the ViewFormer~\cite{li2025viewformer} head as our baseline for occupancy and flow prediction task, and the DiffusionDrive~\cite{liao2024diffusiondrive} head for end-to-end driving task. The adapter and FSA are configured with 4 and 3 attention layers, respectively.

\noindent\textbf{Training.} The world model adopts a two-step training strategy, with the world encoder first trained for 40 epochs, followed by the future decoder trained for 24 epochs. Both stages utilize the AdamW optimizer ~\cite{loshchilov2018decoupled} and are conducted on 32 NVIDIA H20 GPUs with a total batch size of 32. In the future decoding training phase, we predict states at 0.5-second intervals over a 3-second horizon. The loss weights $\omega_l$ and $\omega_s$ in Eq.(\ref{eq:dynamic_loss}) are set to 1.0 and 0.5, respectively. For ray sampling, we first utilize DetZero~\cite{ma2023detzero} to identify moving objects and aggregate point clouds from the time range [-2s, +2.5s] to reconstruct the scene. We then select up to 8000 rays from LiDAR view through voxel-based downsampling and 10000 rays from image view based on pixel semantics. Additionally, for dynamic-aware ray sampling,  we sample up to 2,000 rays within the bounding boxes of moving objects. For downstream tasks, we follow the official training protocols while configuring the world model to predict 2 future frames for 3D occupancy and flow prediction, and 3 frames for end-to-end driving.

\subsection{Main Results}
In this section, we first evaluate our DriveX model in point cloud forecasting task to demonstrate its superior capability in future forecasting. We then benchmark its effectiveness in improving the performance of occupancy prediction, occupancy flow estimation, as well as end-to-end driving.

\noindent\textbf{Point Cloud Forecasting.} Following \cite{yang2024driving, zhou2025hermes}, we evaluate on points within the range of [-51.2m, 51.2m] on both X and Y axes. As presented in Table \ref{tab:point}, our DriveX model outperforms the previous state-of-the-art method HERMES~\cite{zhou2025hermes} across all timestamps, achieving a reduction of 0.12$\text{m}^2$ and 0.09$\text{m}^2$ in Chamfer Distance at 1.0s and 2.0s horizons. 

\noindent\textbf{Occupancy and Flow Prediction.} We verify our DriveX model on the 3D occupancy and flow prediction tasks, employing the state-of-the-art ViewFormer \cite{li2025viewformer} as baseline. As shown in Table \ref{tab:occ}, for occupancy prediction, the DriveX model boosts ViewFormer's performance by +0.77 mIoU and +1.15 $\text{IoU}_{geo}$, surpassing all previous methods. Notably, while conventional pretrained-based world models like ViDAR \cite{yang2024driving} show marginal improvements on such a strong baseline, DriveX still delivers substantial gains, which we attribute to its ability to learn a more comprehensive understanding of the world. To further investigate its effectiveness in modeling world dynamics, we add a flow head for occupancy motion estimation. The model achieves an mAVE of 0.385, reducing the error by 6.5\%. The performance gains also remain remarkable for both mIoU (+0.93) and $\text{IoU}_{geo}$ (+1.08). These improvements highlight the capability of the DriveX model in enhancing downstream models via future scene prediction.

\noindent\textbf{End-to-end Driving.} Safely planning for autonomous driving requires accurate anticipation of future consequences. By leveraging a world model, we can comprehensively evaluate possible futures to enhance performance. As shown in Table \ref{tab:e2e}, our DriveX model, based on the re-implemented DiffusionDrive-C \cite{liao2024diffusiondrive}, boosts the overall PDMS to 84.5, achieving state-of-the-art performance among camera-only end-to-end driving models.

\subsection{Ablation Study}
In this section, we conduct a series of ablation studies on nuScenes~\cite{caesar2020nuscenes} dataset to investigate the individual components of our DriveX framework. By default, we use an input image resolution of $256 \times 704$ for efficient training. All models are trained with a batch size of 32 on 32 GPUs.

\noindent \textbf{Omni scene modeling.} We first ablate the effects of our omni scene modeling module through occupancy and flow prediction task. As presented in Table \ref{tab:ablate-cms}, DriveX captures only geometric information when supervised solely by depth estimation. Adding semantic supervision enriches the model's semantic understanding, achieving a mIoU of 42.53. However, since the semantic labels from auto-labeling inevitably suffer from information loss, incorporating color reconstruction preserves additional generalizable features, ultimately improving the mIoU to 43.47.

\noindent \textbf{Effect of prediction horizon.} Table \ref{tab:ablate-time} verifies that the downstream performance gain stems from DriveX's predictive capability. As indicated in the $2^{nd}$ row, DriveX achieves significant improvements of +0.9 in mIoU and -0.03 in mAVE compared to the baseline without future prediction. While extending the prediction horizon to 4 frames brings further gains, we adopt 2 as the default setting to balance performance and computational cost.

\noindent \textbf{Decoupled training strategy.} We ablate the effects of our decouple training strategy in Table \ref{tab:ablate-decluple}. For a fair comparison, we train a joint training variant of our DriveX model for 64 epochs, matching the total training duration of the decoupled version. As shown in $1^{st}$ and $2^{nd}$ rows, our decoupled approach significantly improves prediction accuracy, reducing the Chamfer Distance by 0.44$\text{m}^2$ for 3.0s predictions. This validates that separating the world model into world encoding and latent future predicting leads to a better modeling of world evolution. Moreover, the integration of dynamic-aware ray sampling strategy enables the model to focus on motion-critical regions, further reducing the average Chamfer Distance to 1.02$\text{m}^2$. Additionally, we compare direct prediction with autoregressive strategies for future decoding in Table \ref{tab:ablate-future}, where direct prediction demonstrates superior performance, diminishing the average Chamfer Distance from 1.32$\text{m}^2$ to 1.02$\text{m}^2$.

\begin{table}
\centering
\resizebox{\columnwidth}{!}{
\begin{tabular}{ccc|ccc}
\toprule
Sup-D & Sup-S & Sup-C & \makecell[c]{\colorbox[gray]{0.85}{mIoU}} & \makecell[c]{$\text{IoU}_{geo}$} & \makecell[c]{\colorbox[gray]{0.85}{$\text{mAVE}\downarrow$}}\\
\midrule
\cmark & & & 3.96& 59.7& 1.388 \\
\cmark & \cmark& & 42.53& 73.1& 0.396 \\
\cmark &\cmark & \cmark& \textbf{43.47}& \textbf{73.44}& \textbf{0.385} \\
\bottomrule
\end{tabular}
}
\vspace{-9pt}
\caption{Ablation study on components of omni scene modeling module. ``Sup-D'', ``Sup-S'' and ``Sup-C'' denote supervision from depth estimation, semantic segmentation, and color reconstruction, respectively.}
\label{tab:ablate-cms}
\vspace{-9pt}
\end{table}

\begin{table}
\centering
\setlength{\tabcolsep}{1.2em}  
\resizebox{0.98\columnwidth}{!}{
\begin{tabular}{c|ccc}
\toprule
Future frames  & \makecell[c]{\colorbox[gray]{0.85}{mIoU}} & \makecell[c]{$\text{IoU}_{geo}$} & \makecell[c]{\colorbox[gray]{0.85}{$\text{mAVE}\downarrow$}}\\
\midrule
0 & 42.57& 72.60& 0.415 \\
2 & 43.47& \textbf{73.44}& \textbf{0.385} \\
4 & \textbf{43.59}& 73.42& 0.387 \\
\bottomrule
\end{tabular}
}
\vspace{-9pt}
\caption{Effects of prediction horizon in FSA .}
\label{tab:ablate-time}
\vspace{-8pt}
\end{table}

\begin{table}
\centering
\resizebox{\columnwidth}{!}{
\begin{tabular}{ccc|c@{\hspace{10pt}}c@{\hspace{10pt}}c@{\hspace{10pt}}c@{\hspace{10pt}}}
\toprule
\multirow{2}{*}{Joint} & \multirow{2}{*}{Decoupled} & \multirow{2}{*}{D-sample} & \multicolumn{4}{c}{Chamfer Distance ($\text{m}^2$)  $\downarrow$}  \\
& &  & 1.0s & 2.0s & 3.0s& Avg. \\
\midrule
\cmark &  &  & 1.09 & 1.41 & 1.75 &  1.41   \\
 & \cmark &  & 0.83 & 1.03 & 1.31 & 1.06  \\
& \cmark & \cmark & \textbf{0.80} & \textbf{0.99} & \textbf{1.28} & \textbf{1.02}  \\
\bottomrule
\end{tabular}
}
\vspace{-9pt}
\caption{Ablation study on decoupled training strategy. ``D-Sample'' refers dynamic-aware ray sampling. }
\label{tab:ablate-decluple}
\vspace{-10pt}
\end{table}

\begin{table}
\centering
\setlength{\tabcolsep}{8.5pt}  
\resizebox{0.97\columnwidth}{!}{
\begin{tabular}{c|cccc}
\toprule
\multirow{2}{*}{Strategy}  & \multicolumn{4}{c}{Chamfer Distance ($\text{m}^2$) $\downarrow$}  \\
&  1.0s & 2.0s & 3.0s & Avg.   \\
\midrule
Autoregressive-pred & 0.99 & 1.21 & 1.76 & 1.32   \\
Direct-pred & \textbf{0.80} & \textbf{0.99} & \textbf{1.28} &  \textbf{1.02} \\
\bottomrule
\end{tabular}
}
\vspace{-9pt}
\caption{Comparison of autoregressive and direct prediction strategies for future state prediction.}
\vspace{-12pt}
\label{tab:ablate-future}
\end{table}

\begin{table}
\centering
\resizebox{0.98\columnwidth}{!}{
\begin{tabular}{c|c@{\hspace{15pt}}c@{\hspace{12pt}}c@{\hspace{12pt}}c@{\hspace{12pt}}|c}
\toprule
\multirow{2}{*}{Dataset fraction} & \multicolumn{4}{c|}{Chamfer Distance ($\text{m}^2$)  $\downarrow$} &  \multirow{2}{*}{Time}\\
  & 1.0s & 2.0s & 3.0s& Avg. & \\
\midrule
  10\% & 2.08 & 2.30 & 2.51 & 2.30  & 1.2h\\
  50\% & 1.36 & 1.53 & 1.70 & 1.53  & 5.7h\\
  100\% & \textbf{0.80} & \textbf{0.99} & \textbf{1.28} & \textbf{1.02} & 11.2h\\
\bottomrule
\end{tabular}
}
\vspace{-9pt}
\caption{Ablation study on data scale. All experiments are conducted with the same training epochs.}
\label{tab:ablate-data}
\vspace{-14pt}
\end{table}

\noindent \textbf{Data Scaling.} A significant promise of self-supervised world models lies in their ability to achieve consistent performance improvements with increasing data scale. We conduct this study on varying fractions of NuScenes training set. Table \ref{tab:ablate-data} demonstrates consistent performance gains across all metrics as training data increases. When scaling from 50\% to 100\% of the dataset, the model still achieves substantial gains, reducing Chamfer Distance by 0.51$\text{m}^2$.

\subsection{Efficiency and Robustness Analysis} DriveX employs a shared world encoder across tasks with minimal computational overhead, adding only 48ms latency and 12.8M parameters for occupancy and flow prediction task over ViewFormer~\cite{li2025viewformer} (150ms, 103.8M). Similarly, for end-to-end driving task, it introduces merely
42ms latency and 11.9M parameters over DriffusionDrive-C~\cite{liao2024diffusiondrive} (90ms, 56.0M).  Both experiments are conducted on an H20 GPU. Additionally, we evaluate DriveX’s robustness under different weather and lighting conditions in Table~\ref{tab:roboust}. DriveX achieves consistent performance gains across varying environments, confirming its practical reliability.

\begin{table}[]
    \centering
    \resizebox{0.95\linewidth}{!}{
    \setlength{\tabcolsep}{9pt}
    \begin{tabular}{l|cc|cc}
    \hline
    {Method}& {Day} & {Night} & {Sunny} &  Rainy\\
    \hline
    ViewFormer & 43.42&  28.57 & 42.28 & 44.14\\
    + DriveX-S & \textbf{44.26} & \textbf{29.61} & \textbf{43.25} & \textbf{44.93} \\
    \rowcolor{lightblue}
    \textit{Improvement}& \textbf{\textit{+0.84}} & \textbf{\textit{+1.04}} & \textbf{\textit{+0.97}} & \textbf{\textit{+0.79}} \\
    \hline
    \end{tabular}
    }
    \vspace{-11pt}
    \caption{Occupancy prediction performance (mIoU) under different lighting and weather conditions on nuScenes validation set.}
    \label{tab:roboust}
    \vspace{-10pt}
\end{table}

\section{Conclusion}
In this paper, we present DriveX, a self-supervised world model that learns generalizable scene dynamics and holistic representations from driving videos. DriveX introduces Omni Scene Modeling, unifying multimodal supervision for comprehensive scene understanding, and employs a decoupled latent modeling strategy to simplify the learning of complex dynamics. Additionally, we develop the Future Spatial Attention paradigm to enhance downstream task adaptability. Extensive experiments on the nuScenes and NAVSIM datasets demonstrate the effectiveness of our approach, paving the way for developing more practical and scalable world models in autonomous driving systems.


{
    \small
    \bibliographystyle{ieeenat_fullname}

}


\end{document}